\documentclass{article}
\usepackage{spconf,amsmath,graphicx,hyperref}
\usepackage{multirow} 
\usepackage{pifont}
\newcommand{\cmark}{\ding{51}}
\newcommand{\xmark}{\ding{55}}

\title{Beyond Simple Fusion: Adaptive Gated Fusion for Robust Multimodal Sentiment Analysis}
\name{Han Wu$^{1*}$\thanks{Equal contribution.}, Yanming Sun$^{1*}$, Yunhe Yang$^2$, Derek F. Wong$^1$\sthanks{Corresponding author.}}

\address{
$^1$NLP$^2$CT Lab, University of Macau, $^2$iFLYTEK Co., Ltd.\\
nlp2ct.\{wuhan, yanming\}@gmail.com, yhyang29@iflytek.com, derekfw@um.edu.mo }

\begin{document}
\ninept
\maketitle
\begin{abstract}
Multimodal sentiment analysis (MSA) leverages information fusion from diverse modalities (e.g., text, audio, visual) to enhance sentiment prediction. However, simple fusion techniques often fail to account for variations in modality quality, such as those that are noisy, missing, or semantically conflicting. This oversight leads to suboptimal performance, especially in discerning subtle emotional nuances. To mitigate this limitation, we introduce a simple yet efficient \textbf{A}daptive \textbf{G}ated \textbf{F}usion \textbf{N}etwork that adaptively adjusts feature weights via a dual gate fusion mechanism based on information entropy and modality importance. This mechanism mitigates the influence of noisy modalities and prioritizes informative cues following unimodal encoding and cross-modal interaction. Experiments on CMU-MOSI and  CMU-MOSEI show that AGFN significantly outperforms strong baselines in accuracy, effectively discerning subtle emotions with robust performance. Visualization analysis of feature representations demonstrates that AGFN enhances generalization by learning from a broader feature distribution, achieved by reducing the correlation between feature location and prediction error, thereby decreasing reliance on specific locations and creating more robust multimodal feature representations.
\end{abstract}
\section{Introduction}
\label{sec:intro}
\begin{figure}[t]
    \centering
    \includegraphics[width=0.95\linewidth]{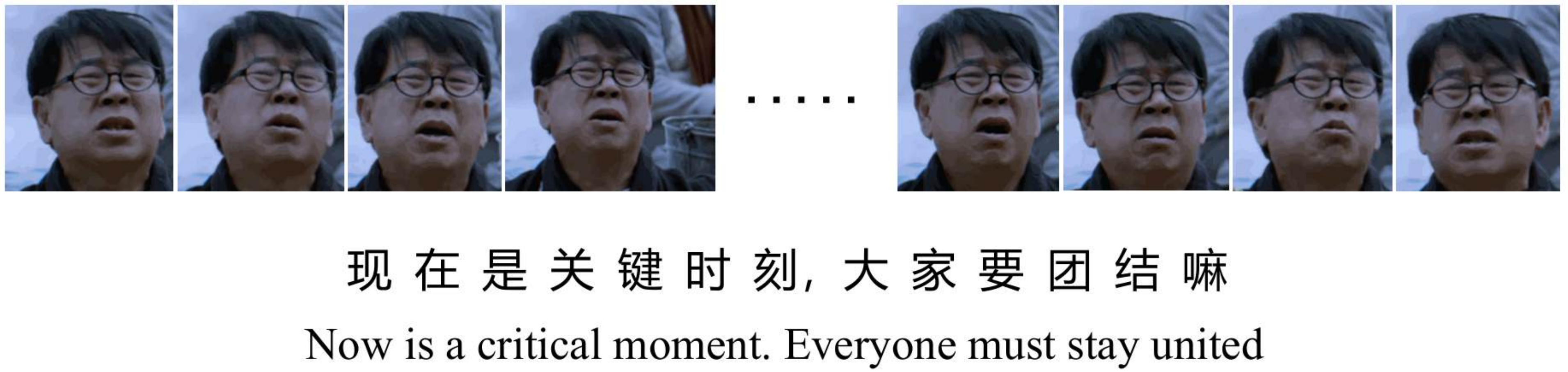}
    \caption{An example of modal inconsistency from the CH-SIMS dataset \cite{yu2020ch}. In this case, the text promotes unity, but the visuals convey a negative affect, creating a conflict.}
    \label{fig:modality_inconsistencies}
\end{figure}

Multimodal sentiment analysis (MSA) promises richer emotion understanding by fusing text, audio, and visual cues, yet in practice, it often struggles to capture the full complexity of human emotions. Consider Figure \ref{fig:modality_inconsistencies}: in nuanced scenarios like sarcasm, textual sentiment might be positive while vocal or visual cues convey the true negative emotion. Worse still, real-world data often contain noisy or irrelevant modalities, such as background chatter or non-informative facial expressions, which further degrade performance if not suppressed. Despite advances in pre-trained models \cite{kim2023aobert,wang2023tetfn,ma2023transformer}, most MSA systems still rely on simple fusion mechanisms, implicitly treating all modalities as equally trustworthy, regardless of their actual quality, relevance, or potential for contradiction. This assumption is fundamentally at odds with the importance, heterogeneous nature of human expression \cite{DBLP:journals/corr/BaltrusaitisAM17}.

To better handle noisy, conflicting, or unreliable signals across modalities, we propose an Adaptive Gated Fusion Network (AGFN) for MSA. AGFN employs a dual-gated fusion mechanism that explicitly models information reliability (via information entropy gate) and modality importance (via modality importance gate), enabling the model to suppress misleading cues and amplify trustworthy ones. This allows the model to focus on the most relevant cues, leading to more robust and accurate sentiment prediction. AGFN achieves superior accuracy on CMU-MOSI and CMU-MOSEI, significantly outperforming the baselines. 

Beyond accuracy, we seek to explain why AGFN works — what makes its fusion more robust? To better understand where the performance gains originate, we analyze how AGFN reshapes the learned feature space. Our analysis reveals that the adaptive gated fusion mechanism builds a superior and more robust feature space by significantly reducing Predictive Spatial Correlation (PSC). This reduction indicates a weaker link between a feature’s spatial location and its prediction error, confirming that the model avoids over-relying on spatially fixed or modality-biased cues. Instead, it prioritizes informative modalities while suppressing noise. Consequently, the learned representations achieve greater generalization and enhanced resilience to modality inconsistencies.

\section{Our Method}

\begin{figure*}[t]
    \centering
    \includegraphics[width=1\linewidth]{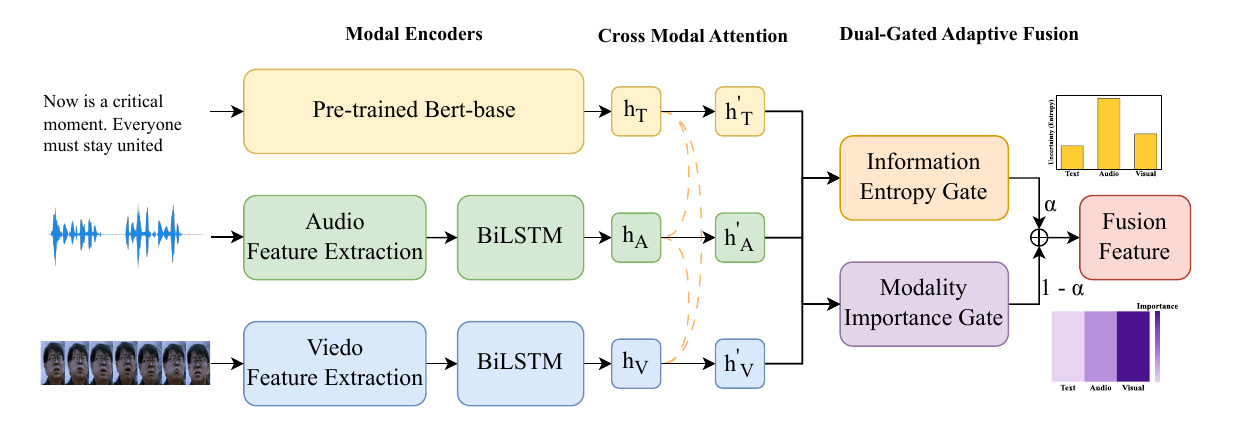}
    \caption{Overview of the AGFN architecture. Text, audio, and video inputs are processed by modality-specific encoders. Cross-modal attention refines these features. The enhanced features are fed into the dual-gated adaptive fusion module: (1) The information entropy gate assesses modality certainty, and (2) the modality importance gate learns sample-specific weights. A learned parameter $\alpha$ adaptively balances contributions from both gates to produce the final fused feature representation.}
    \label{fig:fusion_gate}
\end{figure*}

\subsection{Preliminary}
Multimodal sentiment analysis (MSA) aims to learn a mapping function that predicts a sentiment score $\hat{y}$ based on inputs from multiple modalities, typically textual ($T$), acoustic ($A$), and visual ($V$). The standard deep learning approach involves first processing each modality independently using dedicated unimodal encoders ($E_T, E_A, E_V$) to generate intermediate representations:
\begin{equation}
h_T = E_T(T), \quad h_A = E_A(A), \quad h_V = E_V(V)
\end{equation}Where $h_T, h_A, h_V$ capture modality-specific features.
A crucial step follows where these unimodal representations are integrated through a fusion mechanism to create a joint multimodal representation, $h_{fused}$. The most common baseline method directly concatenates these representations:
\begin{equation}
    h_{\mathrm{fused}} = \mathrm{Concat}(h_T, h_A, h_V)
\end{equation}
This fused representation $h_{\mathrm{fused}}$ is processed by a prediction network, denoted as $f$, to produce the sentiment prediction $\hat{y}$. 
The model is trained by finding the parameters $\Theta^*$ that minimize a chosen loss function $\mathcal{L}$ comparing the predictions $\hat{y}$ to the ground truth labels $y$ across the training dataset $\mathcal{D}$:

\begin{equation}
\Theta^* = \underset{\Theta}{\arg\min} \sum_{(h_{\mathrm{fused}}, y) \in \mathcal{D}} \mathcal{L}\Big(y, \underbrace{f(h_\mathrm{fused}; \theta_p)}_{\hat{y}}\Big)
\end{equation}

\begin{table}[t]
    \centering
    \caption{Method comparison. 2G-Fusion = Dual-Gated Fusion (IEG, MIG); Cross-Attn = Cross-Attention; Repr. = Representation learning; Self-sup. CL = Self-supervised contrastive learning; INF. REL. = information reliability; MOD. IMP. = modality importance.}
    \setlength{\tabcolsep}{2pt}
    \begin{tabular}{|l|c|c|c|c|}
        \hline
        Method & How & INF. REL. & MOD. IMP. \\ \hline\hline
        MulT \cite{tsai2019multimodal} & Cross-Attn & \xmark & \cmark \\
        MISA \cite{hazarika2020misa}&  Repr.~Learn & \xmark & \xmark \\
        SELF-MM \cite{yu2021learning}   & Self-sup.~CL & \xmark & \xmark \\
        AGFN  & 2G-Fusion & \cmark & \cmark \\ \hline
    \end{tabular}
    \label{tab:method_comparison}
\end{table}

\begin{table*}[!t]
\centering
\caption{Performance comparison on the CMU-MOSI and CMU-MOSEI datasets.  All results are averaged over three runs with different random seeds (1111, 11112, 1113) to ensure robustness. Best results are in \textbf{bold}.}
\begin{tabular}{|l||cccc|cccc|} 
\hline
\multirow{2}{*}{Model} & \multicolumn{4}{c|}{CMU-MOSI} & \multicolumn{4}{c|}{CMU-MOSEI} \\ 
\cline{2-9} 
                       & Acc-2$\uparrow$ & F1$\uparrow$ & Acc-7$\uparrow$ & MAE$\downarrow$ & Acc-2$\uparrow$ & F1$\uparrow$ & Acc-7$\uparrow$ & MAE$\downarrow$ \\ 
\hline
TFN \cite{zadeh2017tensor}                   & 76.68 & 76.69 & 35.37  & 96.47  & 74.42 & 75.58  & 50.85  & 57.84  \\ 
LMF \cite{liu2018efficient}                  & 77.02 & 76.97 & 34.94 & 96.86  & 78.43 & 79.15  & 51.15 & 57.56  \\ 
MFN \cite{zadeh2018memory}                   & 76.68  & 76.73 & 34.55 & 97.77  & 78.54 & 79.21  & 51.84 & 57.26  \\ 
MISA \cite{hazarika2020misa}                 & 80.95  & 80.97 & 43.44 & 77.05  & 79.72 & 80.43   & 51.99 & 54.84  \\ 
TETFN \cite{wang2023tetfn}                   & 81.92 & 81.87 & 45.14 & 73.58  & 80.4 & 81.07  & 54.02 & 54.10   \\ 
SELF-MM \cite{yu2021learning}                & 82.56 & 82.55 & 46.36 & 71.65  & 78.11 & 78.93  & 53.14 & \textbf{53.25}  \\ 
\textbf{AGFN}                                & \textbf{82.75} & \textbf{82.68} & \textbf{48.69} & \textbf{71.02}  & \textbf{84.01} & \textbf{84.11}  & \textbf{54.30} & 53.57   \\ 
\hline
\end{tabular}
\label{tab:main_madi_results} 
\end{table*}

\begin{table}[t]
    \centering
    \caption{Ablation study on the gating design of AGFN on the CMU-MOSI dataset. IEG = Information Entropy Gate; MIG = modality importance gate; GFM = Gating Fusion Mechanism.}
    \begin{tabular}{|l||cccc|}
        \hline
        Model & Acc-2 $\uparrow$ & F1 $\uparrow$ & Acc-7 $\uparrow$ & MAE $\downarrow$ \\ \hline\hline
        AGFN & \textbf{82.75} & \textbf{82.68} & \textbf{48.69} & \textbf{71.02} \\ \hline
        AGFN w/ IEG & 81.95 & 81.91 & 45.36 & 73.13 \\
        AGFN w/ MIG & 82.56 & 82.52 & 46.35 & 72.53 \\ 
        AGFN w/o GFM & 82.46 & 82.45 & 46.65 & 72.31 \\ \hline
    \end{tabular}
    \label{tab:ablation_focused}
\end{table}

\subsection{Adaptive Gated Fusion Network (AGFN)}
Effective multimodal fusion must account for both information reliability and modality importance, because relying on only one aspect leads to suboptimal performance: models may either suppress informative but uncertain signals, or amplify noisy but salient features. As shown in Table~\ref{tab:method_comparison}, prior methods typically address only one of these dimensions. To bridge this gap, AGFN introduces a dual-gated fusion mechanism that explicitly models both.

First, the \textbf{information entropy gate} measures the reliability of each modality via its feature entropy $H(h_m)$. Lower entropy implies higher certainty and thus a larger weight in fusion. Given projected logits $z_m$ from concatenated features, the entropy-gated fusion is:
\begin{equation}
h_{\text{entropy}} = \sum_{m\in\{T,A,V\}} \mathrm{Softmax}_m\!\left(z_m\, e^{-H(h_m)/\tau}\right) h_m
\end{equation}

Meanwhile, the \textbf{modality importance gate} captures sample-specific modality importance by generating gating factors $\sigma(z)$ from the concatenated features and re-scaling each modality before projection:
\begin{equation}
h_{\text{importance}} = W_f\!\left[\sigma(z)\odot h_T,~\sigma(z)\odot h_A,~\sigma(z)\odot h_V\right]
\end{equation}

A learnable coefficient $\alpha\in[0,1]$ then balances the two gated results to form the final representation:
\begin{equation}\label{eq:madi_fusion}
h_{\mathrm{fused}} = \alpha\, h_{\mathrm{entropy}} + (1-\alpha)\, h_{\mathrm{importance}}
\end{equation}
This $h_{\mathrm{fused}}$ is used for sentiment prediction.

During training, we minimize the L1 regression loss and apply Virtual Adversarial Training (VAT)~\cite{miyato2018virtual} to enhance robustness. VAT finds adversarial perturbations $h_{\mathrm{fused}}^{\mathrm{adv}}$ and enforces consistency:
\begin{equation}
\mathcal{L}_{\mathrm{total}} = \mathcal{L}_{\mathrm{L1}} + \lambda\,\mathcal{L}_{\mathrm{MSE}}\!\left(f(h_{\mathrm{fused}}^{\mathrm{adv}}), f(h_{\mathrm{fused}})\right)
\end{equation}

\section{Experiments}
\subsection{Experimental Setup}
\subsubsection{Datasets}
We evaluate our method on standard multimodal sentiment analysis (MSA) benchmarks: CMU-MOSI \cite{zadeh2016mosi} (2,200 movie review clips) and the larger CMU-MOSEI \cite{zadeh2018multimodal} (23,000 diverse clips). Both datasets contain video clips annotated with text, audio, and visual modalities, and sentiment labels ranging from -3 to +3. For qualitative analysis, we additionally include the CH-SIMS dataset \cite{yu2020ch}, which uniquely provides both multimodal and independent unimodal sentiment annotations. All datasets follow a standard 70\% training, 10\% validation, and 20\% testing split.

\subsubsection{Evaluation Metrics}
We evaluate performance using four standard MSA metrics. Acc-2 measures binary classification accuracy for positive versus non-positive sentiment. F1-score provides the harmonic mean of precision and recall for this binary task. Acc-7 evaluates accuracy across seven sentiment classes, based on integer-rounded predictions on the [-3, +3] scale. MAE quantifies the average absolute deviation between predicted continuous scores and ground-truth values.

\subsubsection{Models and Hyperparameters}
All models are implemented in PyTorch using the MMSA framework \cite{yu2021learning} and are trained on NVIDIA V100 GPUs. Textual features are extracted with BERT-base \cite{devlin2018bert}; audio and visual features are obtained via COVAREP \cite{degottex2014covarep} and FACET\footnote{\url{https://github.com/BCG-X-Official/facet}}, respectively. We optimize with AdamW \cite{DBLP:conf/iclr/LoshchilovH19}: BERT fine-tuning uses lr=5e-5, other components use lr=1e-4, both decayed via cosine annealing to 1e-6. Gradient clipping (max norm=1.0) and VAT regularization (weight=0.1) are applied. For CMU-MOSI: batch size=32, weight decay=0.01, VAT steps=5, BiLSTM layers=3, early stopping patience=8 (based on validation MAE). For CMU-MOSEI: batch size=128, weight decay=0.05, VAT steps=3, BiLSTM layers=2, early stopping patience=4 (based on validation MAE).

\subsection{Main Results}
As shown in Table \ref{tab:main_madi_results}, AGFN achieves state-of-the-art or highly competitive performance on both CMU-MOSI and CMU-MOSEI. On CMU-MOSI, it obtains the best results across all metrics: Acc-2 (82.75\%), F1 (82.68\%), Acc-7 (48.69\%), and MAE (71.02), surpassing strong baselines like SELF-MM. On the larger CMU-MOSEI, AGFN leads in Acc-2 (84.01\%), F1 (84.11\%), and Acc-7 (54.3\%), while achieving the second-best MAE (53.57), narrowly behind SELF-MM. Binary predictions are classified as `negative' if score $<$ 0, `non-negative' otherwise. These results demonstrate AGFN’s consistent effectiveness in multimodal sentiment analysis across diverse benchmarks.

\subsection{Ablation Study}
\subsubsection{Effect of Information Entropy Gate (IEG)} As shown in Table \ref{tab:ablation_focused}, we evaluate the impact of removing the Information Entropy Gate (IEG), a component designed to regulate feature fusion based on the information entropy of input modalities. Without IEG, the model exhibits clear performance degradation: Acc-2 drops from 82.75 to 81.95, Acc-7 drops from 48.69 to 45.36, and MAE increases from 71.02 to 73.13. These results confirm that IEG plays a crucial role in enhancing the model's ability to discern subtle differences in sentiment intensity by effectively leveraging modality-specific uncertainty information during the fusion process.

\subsubsection{Effect of Modality Importance Gate (MIG)} As shown in Table~\ref{tab:ablation_focused}, we evaluate the effect of removing the Modality Importance Gate (MIG), a component that adaptively weights and integrates multimodal features based on modality importance. Its removal leads to mild but consistent performance degradation, especially in fine-grained metrics: Acc-7 drops from 48.69 to 46.35, and MAE increases from 71.02 to 72.53. Binary metrics also slightly decline, with Acc-2 falling from 82.75 to 82.56 and F1 from 82.68 to 82.52, indicating a limited yet non-negligible impact. These results suggest MIG is not a dominant contributor, but provides useful auxiliary guidance by helping the model focus on more reliable modalities, yielding incremental gains particularly in regression-oriented sentiment estimation.

\subsubsection{Effect of Adaptive Gating Fusion Mechanism (GFM)} As shown in Table \ref{tab:ablation_focused}, the full adaptive gating fusion achieves the best results. Using only IEG suppresses unreliable modalities but over-prunes informative cues, hurting fine-grained metrics (lower Acc-7, higher MAE). Using only MIG keeps coarse performance competitive but is less stable under cross-modal conflicts, also degrading fine-grained accuracy. This supports a two-step mechanism without contextual cues: IEG enforces reliability by down-weighting high-entropy signals, while MIG amplifies salient features at the instance level; together they provide reliability-aware, sample-adaptive weighting that improves both coarse (Acc-2/F1) and fine-grained (Acc-7/MAE) prediction.

\section{Analysis}

\subsection{Feature Representations Visualization}
To understand the strengths of our proposed adaptive fusion model (AGFN), we perform a comparative visualization analysis. Specifically, we examine the final fused feature representations obtained by the simple fusion baseline and our AGFN model on the CMU-MOSI test set using t-SNE projections (perplexity = 30, learning rate = 200, early exaggeration = 12, n\_iter = 2000, init = 'pca') \cite{maaten2008visualizing}. To quantitatively evaluate the quality of the feature spaces, we introduce the Prediction Space Correlation (PSC) metric, defined as:

\begin{equation}
\label{eq:psc}
\text{PSC} = \frac{\lvert \operatorname{corr}(x_{\text{coords}}, \text{errors}) \rvert + \lvert \operatorname{corr}(y_{\text{coords}}, \text{errors}) \rvert}{2}
\end{equation}
This metric computes the average of the absolute Pearson correlation coefficients between each coordinate axis in the two-dimensional feature space and the model’s prediction errors. A lower PSC value indicates a weaker correlation between feature space positions and prediction errors, implying higher quality and more robust feature representations. In addition, we focus on the distribution patterns of high-error samples within the feature space, defining high-error samples as those corresponding to the top 10\% of prediction errors.

\begin{figure}[t]
    \centering
    \begin{minipage}{0.23\textwidth} 
        \centering
        \includegraphics[width=\linewidth]{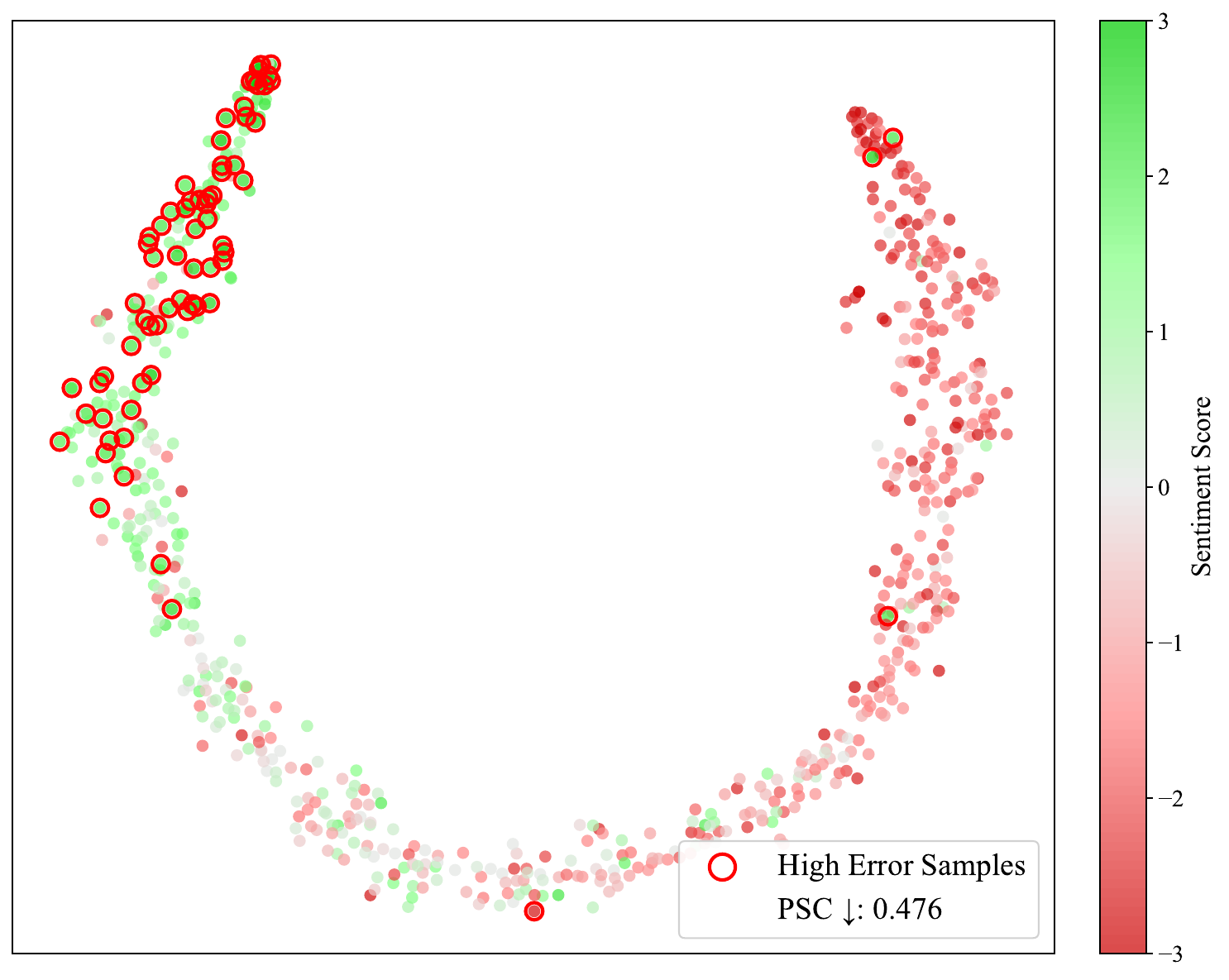} 
    \end{minipage}
    \hspace{0.2cm} 
    \begin{minipage}{0.23\textwidth} 
        \centering
        \includegraphics[width=\linewidth]{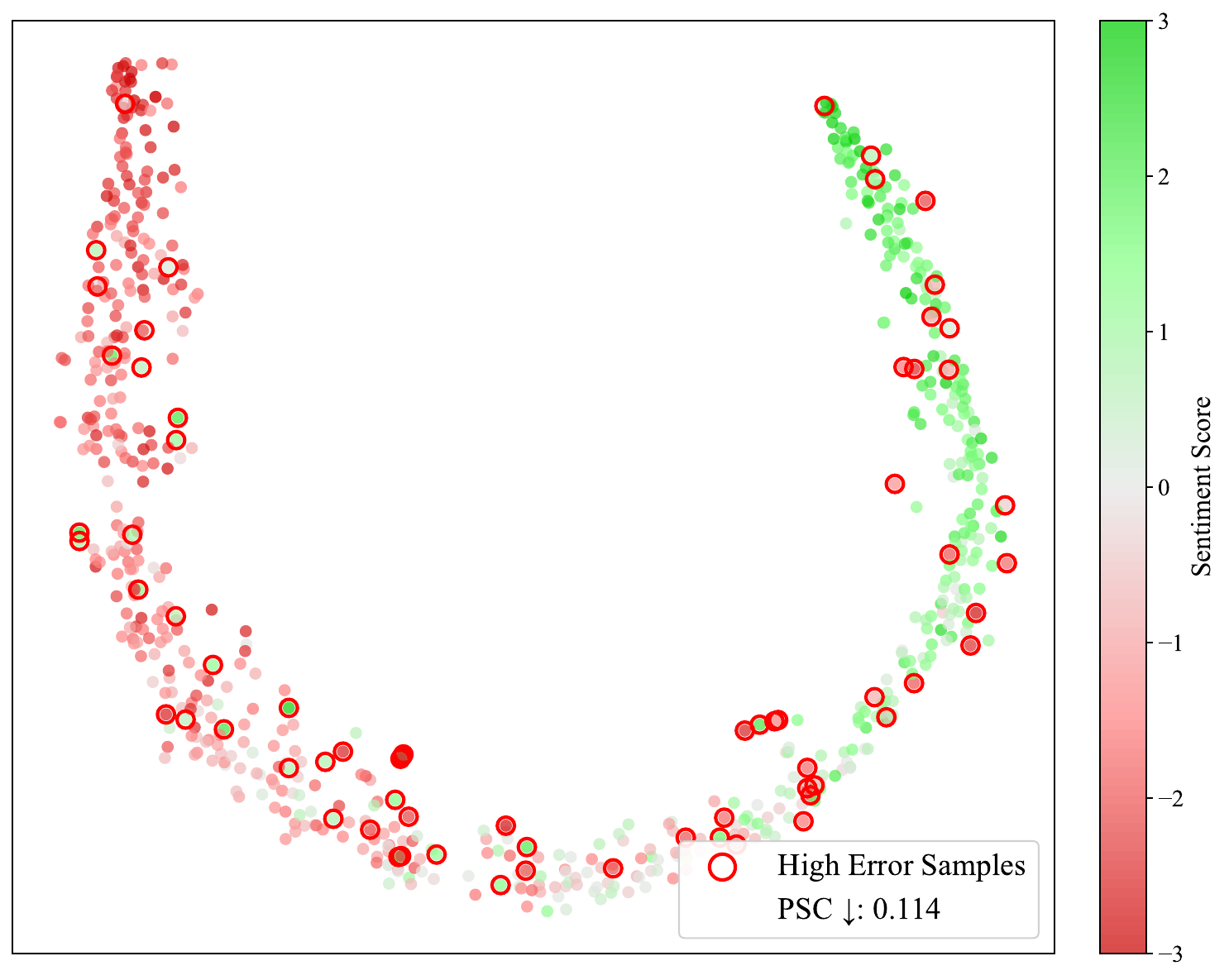}
    \end{minipage}
    
    \caption{t-SNE visualization of final fused representations on the CMU-MOSI test set. \textbf{Left:} Features learned via a simple fusion (concatenation) baseline. \textbf{Right:} Features learned using an adaptive fusion strategy. Points are colored by sentiment polarity.}
    \label{fig:tsne_comparison} 
\end{figure}

The adaptive fusion method learns more robust feature representations by reducing the correlation between feature location and prediction error. As shown in Figure 3, both methods exhibit a smooth sentiment gradient from red (negative) to green (positive). However, the simple fusion baseline (left) concentrates high-error samples in specific regions, indicating systematic failures. In contrast, adaptive fusion (right) distributes errors more uniformly and achieves a significantly lower PSC, confirming its ability to decouple feature geometry from prediction bias.

\subsection{Case Study}
\begin{figure}[t]
    \centering
    \includegraphics[width=1\linewidth]{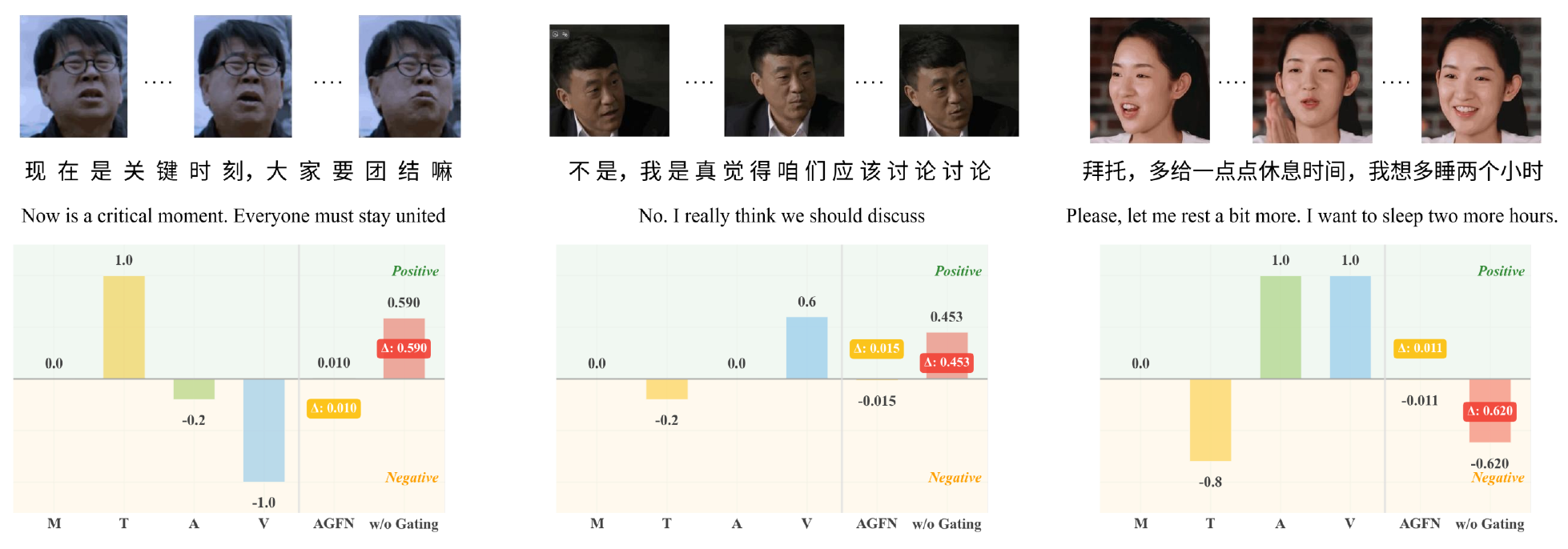}
    \caption{Case Study on the CH-SIMS Dataset. Each case displays the unimodal sentiment scores (T: Text, A: Audio, V: Visual), the ground truth multimodal label (M), and the prediction results from our full AGFN model and the model without gating ``w/o Gating").}
    \label{fig:case_study}
\end{figure}

To qualitatively demonstrate AGFN’s ability to handle modality conflicts, we analyze representative examples from the CH-SIMS dataset \cite{yu2020ch}. As shown in Figure \ref{fig:case_study}, for three instances with a neutral ground-truth label (M: 0.0) but strong cross-modal conflicts, AGFN consistently predicts scores close to neutral. This highlights its effectiveness in resolving conflicts — such as positive text with negative visuals (Case 1), positive visuals with negative text (Case 2), or strongly negative text with positive audio/visual cues (Case 3).

In contrast, the ``w/o Gating" model performs poorly, as its predictions are heavily swayed by a single dominant or conflicting modality, resulting in significant deviations from the neutral label. These results underscore the critical role of AGFN’s adaptive gating mechanism, which dynamically adjusts the influence of each modality based on contextual reliability, thereby mitigating the impact of inconsistencies and yielding robust sentiment predictions.

\section{Related Work}
\subsection{Multimodal Sentiment Analysis and Fusion Strategies}

Multimodal sentiment analysis leverages multiple data streams \cite{das2023multimodal} to interpret sentiment more effectively than unimodal approaches focusing solely on text \cite{li2016text,xu2019chinese} or audio \cite{luo2019audio,garcia2021sentiment}. By integrating complementary cues, MSA can better decipher complex expressions. While progress has been made in developing unimodal feature extractors (e.g., BERT variants \cite{devlin2018bert}, CNNs/ResNets \cite{DBLP:conf/cvpr/HeZRS16}), the central challenge in MSA lies in effectively integrating these features through modality fusion \cite{perez2013utterance,DBLP:journals/corr/BaltrusaitisAM17}. Traditional fusion strategies include early fusion (concatenating features at input \cite{morency2011towards}), which may capture low-level correlations but is sensitive to asynchrony and noise, and late fusion (merging unimodal predictions \cite{nojavanasghari2016deep,tran2015learning}), which offers robustness to input noise and modality asynchrony but overlooks inter-modal importances during learning. Hybrid approaches \cite{wollmer2013youtube} attempt to leverage the advantages of both, but require careful grouping and often lead to increased complexity.

\subsection{Attention-Based Fusion and Current Challenges}

Traditional fusion often fails to capture importance and context-dependent cross-modal relations. Intermediate fusion via attention enables models to reweight features within/across modalities on the fly, as in MulT’s directional cross-modal attention and MISA’s shared/private representations \cite{tsai2019multimodal, hazarika2020misa}. Yet it remains hard to adapt fusion weights to modality reliability under noise or conflicts. We address this with AGFN, which explicitly jointly models information reliability via the Information Entropy Gate (IEG) and modality importance via the modality importance gate (MIG). 

\section{Conclusion}
AGFN is designed to handle real-world modality imperfections, such as noise, missing data, and semantic conflicts, that degrade performance in multimodal sentiment analysis. AGFN’s core innovation is a dual-gated fusion mechanism that importantly weights modalities by jointly modeling information reliability (via information entropy gate) and modality importance (via modality importance gate). This enables AGFN to suppress noisy or conflicting signals, such as sarcasm, and prioritize informative cues. Experiments on CMU-MOSI and CMU-MOSEI show AGFN achieves state-of-the-art performance. Crucially, visualization and our PSC metric confirm that AGFN learns a more robust feature space by decoupling prediction accuracy from fixed feature locations, a fundamental advancement over naive fusion methods.

\vfill\pagebreak

\bibliographystyle{IEEEbib}

\begin{thebibliography}{10}

\bibitem{kim2023aobert}
Kyeonghun Kim and Sanghyun Park,
\newblock ``Aobert: All-modalities-in-one bert for multimodal sentiment analysis,''
\newblock {\em Information Fusion}, 2023.

\bibitem{wang2023tetfn}
Di~Wang, Xutong Guo, Yumin Tian, Jinhui Liu, LiHuo He, and Xuemei Luo,
\newblock ``Tetfn: A text enhanced transformer fusion network for multimodal sentiment analysis,''
\newblock {\em Pattern Recognition}, 2023.

\bibitem{ma2023transformer}
Hui Ma, Jian Wang, Hongfei Lin, Bo~Zhang, Yijia Zhang, and Bo~Xu,
\newblock ``A transformer-based model with self-distillation for multimodal emotion recognition in conversations,''
\newblock {\em IEEE Transactions on Multimedia}, 2023.

\bibitem{DBLP:journals/corr/BaltrusaitisAM17}
Tadas Baltrusaitis, Chaitanya Ahuja, and Louis{-}Philippe Morency,
\newblock ``Multimodal machine learning: {A} survey and taxonomy,''
\newblock {\em CoRR}, 2017.

\bibitem{yu2020ch}
Wenmeng Yu, Hua Xu, Fanyang Meng, Yilin Zhu, Yixiao Ma, Jiele Wu, Jiyun Zou, and Kaicheng Yang,
\newblock ``Ch-sims: A chinese multimodal sentiment analysis dataset with fine-grained annotation of modality,''
\newblock in {\em Proceedings of the 58th annual meeting of the association for computational linguistics}, 2020.

\bibitem{tsai2019multimodal}
Yao-Hung~Hubert Tsai, Shaojie Bai, Paul~Pu Liang, J~Zico Kolter, Louis-Philippe Morency, and Ruslan Salakhutdinov,
\newblock ``Multimodal transformer for unaligned multimodal language sequences,''
\newblock in {\em Proceedings of the conference. Association for computational linguistics. Meeting}. NIH Public Access, 2019.

\bibitem{hazarika2020misa}
Devamanyu Hazarika, Roger Zimmermann, and Soujanya Poria,
\newblock ``Misa: Modality-invariant and-specific representations for multimodal sentiment analysis,''
\newblock in {\em Proceedings of the 28th ACM international conference on multimedia}, 2020.

\bibitem{yu2021learning}
Wenmeng Yu, Hua Xu, Ziqi Yuan, and Jiele Wu,
\newblock ``Learning modality-specific representations with self-supervised multi-task learning for multimodal sentiment analysis,''
\newblock in {\em Proceedings of the AAAI conference on artificial intelligence}, 2021.

\bibitem{zadeh2017tensor}
Amir Zadeh, Minghai Chen, Soujanya Poria, Erik Cambria, and Louis-Philippe Morency,
\newblock ``Tensor fusion network for multimodal sentiment analysis,''
\newblock {\em arXiv preprint arXiv:1707.07250}, 2017.

\bibitem{liu2018efficient}
Zhun Liu, Ying Shen, Varun~Bharadhwaj Lakshminarasimhan, Paul~Pu Liang, Amir Zadeh, and Louis-Philippe Morency,
\newblock ``Efficient low-rank multimodal fusion with modality-specific factors,''
\newblock {\em arXiv preprint arXiv:1806.00064}, 2018.

\bibitem{zadeh2018memory}
Amir Zadeh, Paul~Pu Liang, Navonil Mazumder, Soujanya Poria, Erik Cambria, and Louis-Philippe Morency,
\newblock ``Memory fusion network for multi-view sequential learning,''
\newblock in {\em Proceedings of the AAAI conference on artificial intelligence}, 2018.

\bibitem{miyato2018virtual}
Takeru Miyato, Shin-ichi Maeda, Masanori Koyama, and Shin Ishii,
\newblock ``Virtual adversarial training: a regularization method for supervised and semi-supervised learning,''
\newblock {\em IEEE transactions on pattern analysis and machine intelligence}, 2018.

\bibitem{zadeh2016mosi}
Amir Zadeh, Rowan Zellers, Eli Pincus, and Louis-Philippe Morency,
\newblock ``Mosi: multimodal corpus of sentiment intensity and subjectivity analysis in online opinion videos,''
\newblock {\em arXiv preprint arXiv:1606.06259}, 2016.

\bibitem{zadeh2018multimodal}
AmirAli~Bagher Zadeh, Paul~Pu Liang, Soujanya Poria, Erik Cambria, and Louis-Philippe Morency,
\newblock ``Multimodal language analysis in the wild: Cmu-mosei dataset and interpretable dynamic fusion graph,''
\newblock in {\em Proceedings of the 56th Annual Meeting of the Association for Computational Linguistics (Volume 1: Long Papers)}, 2018.

\bibitem{devlin2018bert}
Jacob Devlin,
\newblock ``Bert: Pre-training of deep bidirectional transformers for language understanding,''
\newblock {\em arXiv preprint arXiv:1810.04805}, 2018.

\bibitem{degottex2014covarep}
Gilles Degottex, John Kane, Thomas Drugman, Tuomo Raitio, and Stefan Scherer,
\newblock ``Covarep—a collaborative voice analysis repository for speech technologies,''
\newblock IEEE, 2014.

\bibitem{DBLP:conf/iclr/LoshchilovH19}
Ilya Loshchilov and Frank Hutter,
\newblock ``Decoupled weight decay regularization,''
\newblock in {\em 7th International Conference on Learning Representations, {ICLR} 2019, New Orleans, LA, USA, May 6-9, 2019}. 2019, OpenReview.net.

\bibitem{maaten2008visualizing}
Laurens van~der Maaten and Geoffrey Hinton,
\newblock ``Visualizing data using t-sne,''
\newblock {\em Journal of machine learning research}, 2008.

\bibitem{das2023multimodal}
Ringki Das and Thoudam~Doren Singh,
\newblock ``Multimodal sentiment analysis: a survey of methods, trends, and challenges,''
\newblock {\em ACM Computing Surveys}, 2023.

\bibitem{li2016text}
Dan Li and Jiang Qian,
\newblock ``Text sentiment analysis based on long short-term memory,''
\newblock in {\em 2016 First IEEE International Conference on Computer Communication and the Internet (ICCCI)}. IEEE, 2016.

\bibitem{xu2019chinese}
Guixian Xu, Ziheng Yu, Haishen Yao, Fan Li, Yueting Meng, and Xu~Wu,
\newblock ``Chinese text sentiment analysis based on extended sentiment dictionary,''
\newblock {\em IEEE access}, 2019.

\bibitem{luo2019audio}
Ziqian Luo, Hua Xu, and Feiyang Chen,
\newblock ``Audio sentiment analysis by heterogeneous signal features learned from utterance-based parallel neural network.,''
\newblock .

\bibitem{garcia2021sentiment}
Mar{\'\i}a~Teresa Garc{\'\i}a-Ord{\'a}s, H{\'e}ctor Alaiz-Moret{\'o}n, Jos{\'e}~Alberto Ben{\'\i}tez-Andrades, Isa{\'\i}as Garc{\'\i}a-Rodr{\'\i}guez, Oscar Garc{\'\i}a-Olalla, and Carmen Benavides,
\newblock ``Sentiment analysis in non-fixed length audios using a fully convolutional neural network,''
\newblock 2021.

\bibitem{DBLP:conf/cvpr/HeZRS16}
Kaiming He, Xiangyu Zhang, Shaoqing Ren, and Jian Sun,
\newblock ``Deep residual learning for image recognition,''
\newblock in {\em 2016 {IEEE} Conference on Computer Vision and Pattern Recognition, {CVPR} 2016, Las Vegas, NV, USA, June 27-30, 2016}. 2016, {IEEE} Computer Society.

\bibitem{perez2013utterance}
Ver{\'o}nica P{\'e}rez-Rosas, Rada Mihalcea, and Louis-Philippe Morency,
\newblock ``Utterance-level multimodal sentiment analysis,''
\newblock 2013.

\bibitem{morency2011towards}
Louis-Philippe Morency, Rada Mihalcea, and Payal Doshi,
\newblock ``Towards multimodal sentiment analysis: Harvesting opinions from the web,''
\newblock in {\em Proceedings of the 13th international conference on multimodal interfaces}, 2011.

\bibitem{nojavanasghari2016deep}
``Deep multimodal fusion for persuasiveness prediction,''
\newblock in {\em Proceedings of the 18th ACM international conference on multimodal interaction}, 2016.

\bibitem{tran2015learning}
Du~Tran, Lubomir Bourdev, Rob Fergus, Lorenzo Torresani, and Manohar Paluri,
\newblock ``Learning spatiotemporal features with 3d convolutional networks,''
\newblock 2015.

\bibitem{wollmer2013youtube}
Martin W{\"o}llmer, Felix Weninger, Tobias Knaup, Bj{\"o}rn Schuller, Congkai Sun, Kenji Sagae, and Louis-Philippe Morency,
\newblock ``Youtube movie reviews: Sentiment analysis in an audio-visual context,''
\newblock {\em IEEE Intelligent Systems}, 2013.

\end{thebibliography}

\end{document}